%% file: sample-sigconf-authordraft.tex
\documentclass[sigconf]{acmart}
\AtBeginDocument{%
  }

\copyrightyear{2025}
\acmYear{2025}
\setcopyright{acmlicensed}
\acmConference[CIKM '25]{Proceedings of the 34th
ACM International Conference on Information and Knowledge
Management}{November 10--14, 2025}{Seoul, Republic of Korea}
\acmBooktitle{Proceedings of the 34th ACM International Conference on
Information and Knowledge Management (CIKM '25), November 10--14, 2025,
Seoul, Republic of Korea}
\acmDOI{10.1145/3746252.3761097}
\acmISBN{979-8-4007-2040-6/2025/11}

\usepackage{multirow}
\usepackage{booktabs}
\usepackage[ruled, linesnumbered]{algorithm2e}
\usepackage{xcolor}
\begin{document}

\newcommand{\algo}[1]{{pFedBayesPT}}
\title{Towards Instance-wise Personalized Federated Learning via Semi-Implicit
Bayesian Prompt Tuning}

\author{Tiandi Ye}
\orcid{0000-0003-0169-457X}
\affiliation{%
 \institution{East China Normal University}
 \city{}
 \state{Shanghai}
 \country{China}
}
\email{tdye@stu.ecnu.edu.cn}

\author{Wenyan Liu}
\orcid{0000-0002-9312-1295}
\affiliation{%
 \institution{Ant Group}
 \city{Hangzhou}
 \state{Zhejiang}
 \country{China}
}
\email{enxuan.lwy@antgroup.com}

\author{Kai Yao}
\orcid{0000-0003-4623-0365}
\affiliation{%
 \institution{Ant Group}
 \city{Hangzhou}
 \state{Zhejiang}
 \country{China}
}
\email{jiumo.yk@antgroup.com}

\author{Lichun Li}
\orcid{0009-0001-7495-1445}
\affiliation{%
 \institution{Ant Group}
 \city{Hangzhou}
 \state{Zhejiang}
 \country{China}
}
\email{lichun.llc@antgroup.com}

\author{Shangchao Su}
\orcid{0000-0001-9925-1114}
\affiliation{%
 \institution{Fudan University}
 \city{}
 \state{Shanghai}
 \country{China}
}
\email{scsu20@fudan.edu.cn}

\author{Cen Chen}
\orcid{0000-0003-0325-1705}
\affiliation{%
 \institution{East China Normal University}
 \city{}
 \state{Shanghai}
 \country{China}
}
\email{cenchen@dase.ecnu.edu.cn}
\authornote{Corresponding author.}

\author{Xiang Li}
\orcid{0000-0003-0945-145X}
\affiliation{%
 \institution{East China Normal University}
 \city{}
 \state{Shanghai}
 \country{China}
}
\email{xiangli@dase.ecnu.edu.cn}

\author{Shan Yin}
\orcid{0009-0000-6630-6051}
\affiliation{%
 \institution{Ant Group}
 \city{Hangzhou}
 \state{Zhejiang}
 \country{China}
}
\email{yinshan.ys@antgroup.com}

\author{Ming Gao}
\orcid{0000-0002-5603-2680}
\affiliation{%
 \institution{East China Normal University}
 \city{}
 \state{Shanghai}
 \country{China}
}
\email{mgao@dase.ecnu.edu.cn}

\renewcommand{\shortauthors}{Tiandi Ye et al.}

\begin{abstract}
  Federated learning (FL) is a privacy-preserving machine learning paradigm that enables collaborative model training across multiple distributed clients without disclosing their raw data. Personalized federated learning (pFL) has gained increasing attention for its ability to address data heterogeneity. However, most existing pFL methods assume that each client's data follows a single distribution and learn one client-level personalized model for each client. This assumption often fails in practice, where a single client may possess data from multiple sources or domains, resulting in significant intra-client heterogeneity and suboptimal performance. To tackle this challenge, we propose \algo~, a fine-grained instance-wise pFL framework based on visual prompt tuning. Specifically, we formulate instance-wise prompt  generation from a Bayesian perspective and model the prompt posterior as an implicit distribution to capture diverse visual semantics. We derive a variational training objective under the semi-implicit variational inference framework. Extensive experiments on benchmark datasets demonstrate that \algo~ consistently outperforms existing pFL methods under both feature and label heterogeneity settings.
\end{abstract}

\ccsdesc[500]{Computing methodologies~Artificial intelligence}


\keywords{Instance-wise Personalization, Federated Learning, Prompt Tuning, Variational Inference}


\maketitle

\input{1_intro}
\input{2_related}
\input{3_preliminary}
\input{4_method}
\input{5_results}

\input{6_conclusion}

\begin{acks}
This work has been supported by the National Natural Science Foundation of China under Grant No. 62202170 and No. 62377012, the Guizhou Provincial Program on Commercialization of Scientific and Technological Achievements (Qiankehezhongyindi [2025] No. 006), the National Social Science Fund of China (Major Program, Grant No. 23\&ZD134), and Ant Group Research Intern Program.
\end{acks}

\newpage
\section{GenAI Usage Disclosure}
We acknowledge that generative AI tools (ChatGPT) were used solely for language editing and improving clarity in the writing of this manuscript. 
All conceptual ideas, experimental designs, analyses, and conclusions are our original work. 
We take full responsibility for the correctness and integrity of the content.

\bibliographystyle{ACM-Reference-Format}
\bibliography{sample-base}

\end{document}

%% file: 1_intro.tex
\section{Introduction}
With the rapid proliferation of artificial intelligence (AI) applications, ranging from personalized recommendation systems to intelligent healthcare, the demand for large-scale and diverse data resources has reached an unprecedented level. Nevertheless, in many real-world scenarios, data are inherently fragmented and locked within organizational silos, such as hospitals, financial institutions, and service providers. Direct data sharing across these entities is often prohibited due to competitive concerns, ethical considerations, or strict regulatory constraints, such as General Data Protection Regulation (GDPR)\footnote{GGPR: https://gdpr-info.eu/} and California Consumer Privacy Act (CCPA)\footnote{CCPA: https://oag.ca.gov/privacy/ccpa}. This situation leads to a paradox: while data abundance exists globally, individual organizations may still face data scarcity for building robust models. Federated learning (FL) has emerged as a compelling paradigm to bridge this gap, enabling collaborative model training across decentralized data sources while preserving privacy and security~\cite{mcmahan2017communication}. By doing so, FL not only unlocks the potential of distributed data but also aligns with the growing societal and legal emphasis on responsible AI development.

The seminal FL algorithm, FedAvg~\cite{mcmahan2017communication}, adopts a simple protocol in which each client independently performs several steps of standard local training, and the server aggregates the updated models by averaging their parameters. FedAvg has shown notable success in various applications, such as keyboard prediction~\cite{hard2018federated} and digital health~\cite{rieke2020future}.
However, its performance degrades significantly under data heterogeneity\footnote{That is, when client data are not independently and identically distributed (non-IID).} due to optimization divergence. 
To alleviate this issue, a number of methods have introduced local regularization techniques to mitigate client drift~\cite{li2020federated,karimireddy2020scaffold,acar2020federated}.
Despite their effectiveness, these approaches still conform to the \textit{generic} FL setting, where all clients share a single global model, thereby limiting the performance of individual clients~\cite{chen2022bridging}.
To overcome this limitation, personalized federated learning (pFL) has emerged as a promising solution. pFL aims to tailor a customized model for each client to better account for data heterogeneity~\cite{tan2022towards}. 
However, most existing pFL methods assume that each client’s data originates from a single distribution, which is not always the case in real-world scenarios. 

In practice, a client’s data may originate from multiple devices or be collected across different times, locations, or environmental conditions, resulting in substantial \textit{intra-client} data heterogeneity~\cite{feng2023towards,chen2019blending}.
For instance, a mobile device may contain both indoor and outdoor images with distinct visual characteristics.
In such cases, the model may exhibit insufficient capacity to represent diverse patterns within a single client. Although increasing model size could enhance expressiveness, it simultaneously heightens the risk of overfitting due to the limited amount of data available on each client. This trade-off often leads to suboptimal performance for existing \textit{client-level} pFL methods~\cite{feng2023towards}, underscoring the need for a more fine-grained, instance-wise personalization strategy that can improve flexibility while mitigating overfitting.

Building on vision Transformer (ViT)~\cite{dosovitskiy2020image} and visual prompt tuning (VPT)~\cite{jia2022visual}, we propose a novel instance-wise pFL framework that generates personalized prompts conditioned on individual image features. 
Given the limited data on clients, excessive local training can easily lead to overfitting. 
To mitigate this, we leverage the regularization capabilities of Bayesian methods and frame prompt tuning as a variational inference problem. 
In particular, to capture rich and diverse visual semantics~\cite{chen2022plot}, we model the posterior distribution of the prompts as an implicit distribution, extending beyond the commonly assumed exponential family distributions. 
In practice, we transform the variational parameters into random variables by randomly masking image features and optimize the model within the framework of semi-implicit variational inference~\cite{yin2018semi}.
This allows us to increase the expressiveness of the model while avoiding additional learnable parameters.

We validate the effectiveness of the proposed approach through comprehensive experiments on DomainNet and CIFAR-100. Our main contributions can be summarized as follows:
\begin{itemize}
\item We propose \algo~, a novel instance-wise personalized federated Bayesian prompt tuning framework. To the best of our knowledge, this is the first attempt to formulate instance-level visual prompt tuning from a Bayesian perspective in the FL setting.
\item We enhance prompt diversity and expressiveness by modeling the prompt posterior as an implicit distribution, and provide a principled variational training objective.
\item Extensive empirical results on benchmark datasets demonstrate that our method outperforms existing pFL baselines.
\end{itemize}

%% file: 2_related.tex
\section{Related Work}
\subsection{Personalized Federated Learning}
FL faces significant challenges when client data are non-IID, which can severely degrade the performance of standard algorithms like FedAvg~\cite{mcmahan2017communication,li2019convergence}. To mitigate this issue, pFL has emerged as a promising solution~\cite{arivazhagan2019federated,collins2021exploiting,oh2021fedbabu,fallah2020personalized,shamsian2021personalized,ghosh2020efficient,sattler2020clustered,ye2023robust,li2021ditto,chen2022bridging,yang2023efficient,feng2023towards,deng2024unlocking}. 
Existing pFL methods can be divided into two primary categories based on the level of personalization: client-wise personalization and instance-wise personalization. Each of these approaches tackles data heterogeneity at different levels, as detailed below. 

\subsubsection{Client-Wise Personalization}
Client-wise personalization methods aim to learn a distinct model for each client to adapt to their specific data distribution. These approaches typically personalize the model at the client level, either by modifying the global model or adapting components of it for each client.
Existing methods can be broadly categorized into four classes: model decoupling~\cite{arivazhagan2019federated,collins2021exploiting,oh2021fedbabu}, clustering~\cite{ghosh2020efficient,sattler2020clustered,ye2023robust}, meta-learning~\cite{fallah2020personalized}, hypernetworks~\cite{shamsian2021personalized,chen2022bridging}, and multi-task learning~\cite{smith2017federated,li2021ditto}, among others. 
Model decoupling methods share feature extractors across clients while allowing task-specific heads to be personalized. For example, FedPer\cite{arivazhagan2019federated} and FedRep\cite{collins2021exploiting} both maintain a shared feature extractor and train a local classifier for each client. In FedPer, the feature extractor and classifier are optimized simultaneously, whereas FedRep first updates the classification head and then fine-tunes the shared extractor to improve generalization. FedBABU~\cite{oh2021fedbabu} further simplifies this approach by training the feature encoder with a randomly initialized classifier and achieving personalization solely through fine-tuning the classification head. 
Clustering-based approaches partition clients into groups with similar data distributions, enabling intra-cluster collaboration while alleviating the effects of inter-cluster heterogeneity. 
Such methods are particularly effective when the overall client population exhibits multiple latent data distributions. For instance, ClusteredFL~\cite{sattler2020clustered} groups clients based on the similarity of their local gradients or model updates, and then trains separate models for each cluster. 
In contrast, IFCA\cite{ghosh2020efficient} performs dynamic clustering during training by assigning each client to the model that best fits its local data, enabling adaptive discovery of latent client groups. 
Meta-learning approaches aim to learn a global initialization that can be rapidly adapted to each client with only a few local updates. 
A representative method, Per-FedAvg~\cite{fallah2020personalized}, extends model-agnostic meta-learning (MAML~\cite{finn2017model}) to the federated setting, training a meta-model that serves as a strong starting point and enables fast personalization on individual clients. 
Hypernetwork-based methods personalize FL models by generating client-specific parameters through a hypernetwork.
For instance, pFedHN~\cite{shamsian2021personalized} jointly trains a hypernetwork and learnable client descriptors to directly generate personalized model weights. A similar idea is also employed in pFedPG~\cite{yang2023efficient}, which leverages a hypernetwork to produce personalized prompts for guiding client-specific adaptation. 
Multi-task learning approaches formulate FL as a joint optimization problem over multiple client-specific objectives.
For example, Ditto~\cite{li2021ditto} jointly optimizes a global model and local personalized models with a regularization term, achieving a balance between fairness, robustness, and personalization.
Similarly, MOCHA\cite{smith2017federated} introduces a principled federated multi-task learning framework that explicitly models task relatedness, emphasizing the trade-off between shared representation and client-specific adaptation.

\subsubsection{Instance-Wise Personalization}
More recently, several works have extended personalization to a finer granularity by exploring instance-wise personalization~\cite{panchal2024flow,feng2023towards}.
For example, Flow~\cite{panchal2024flow} enables each instance to dynamically choose between the local and global model parameters. FedIns~\cite{feng2023towards} supports instance-adaptive inference by dynamically generating the optimal matching SSF~\cite{lian2022scaling} for each instance from a learned SSF pool. 
Such methods represent an important shift from client-level to instance-level personalization, as they allow the model to adapt to intra-client heterogeneity, which is often overlooked in traditional pFL methods. 

In parallel, a growing body of work explores Bayesian approaches for FL~\cite{yurochkin2019bayesian,kotelevskii2022fedpop,zhang2022personalized}, where model parameters are treated as random variables. 
Bayesian methods are attractive for FL as they naturally capture uncertainty in parameter estimation, provide robustness to small local datasets, and help mitigate overfitting. However, most existing works model the entire model in a Bayesian manner, which incurs significant computational overhead. 
Unlike these methods, we model only the \textit{prompt} as a random variable, rather than the entire model. Moreover, we adopt a more expressive posterior distribution through semi-implicit variational inference (SIVI)~\cite{yin2018semi}, which better captures the complexity and diversity of instance-specific prompt distributions than standard Gaussian assumptions.

\subsection{Federated Prompt Tuning}
For image classification tasks, existing federated prompt tuning methods can be broadly categorized into two types: (1) methods designed for pure vision models such as ViT~\cite{li2023visual,jia2022visual,yang2023efficient,feng2023learning,deng2024unlocking}, and (2) methods developed for vision-language pre-trained models such as CLIP~\cite{guo2023promptfl,guo2023pfedprompt,su2024federated,wei2023dual,li2024global}. This paper primarily focuses on the first category.
FedVPT and FedVPT-D~\cite{jia2022visual,yang2023efficient} collaboratively optimize visual prompts using a standard FedAvg procedure. 
Inspired by pFedHN~\cite{shamsian2021personalized}, FedPG~\cite{yang2023efficient} introduces a server-side prompt generator that produces personalized prompts for each client based on their descriptors.
To mitigate catastrophic forgetting caused
by data heterogeneity, FedPR~\cite{feng2023learning} incorporates a null-space projection step into local prompt updates, suppressing harmful gradients that may degrade server-side performance.
More recently, SGPT~\cite{deng2024unlocking} proposes training both shared and group-specific prompts to bridge the gap between global generalization and client-level personalization. 
These approaches highlight that prompt tuning can serve as an efficient alternative to full model finetuning in FL, as prompts require fewer trainable parameters and lower communication overhead, making them well-suited for resource-constrained federated settings.

In parallel, several studies have begun to explore the intersection of prompt tuning and Bayesian optimization~\cite{lu2022prompt,derakhshani2023bayesian,yang2024generating}.
For instance, GPLS~\cite{yang2024generating} investigates modeling prompts as latent variables in the context of continual learning.
In contrast, our method models instance-wise prompts as random variables to explicitly capture intra-client heterogeneity in FL.
Moreover, we assume the posterior over prompts to follow an implicit distribution, rather than a standard Gaussian as in the works~\cite{derakhshani2023bayesian,yang2024generating}, which proves significantly more expressive for modeling complex prompt variations.

%% file: 3_preliminary.tex
\section{Preliminary}
\subsection{Personalized Federated Learning}
In an FL system with $N$ clients, each client $k$ holds a local dataset $\mathcal{D}_k = \{(x_k^j, y_k^j)\}_{j=1}^{|\mathcal{D}_k|}$. 
The goal of \textit{generic} FL is to collaboratively train a single global model by solving the following objective:
\begin{equation}\label{eq:generic_FL}
    \min_{\theta} \{ f_g := \frac{1}{N} \sum_{k=1}^N f_k(\theta) \},
\end{equation}
where $\theta$ denotes the global model parameters, and $f_k(\theta)$ represents the empirical loss on client $k$, defined as:
\begin{equation}
    f_k(\theta) := \frac{1}{|\mathcal{D}_k|} \sum_{(x, y) \in \mathcal{D}_k} \ell((x, y); \theta).     
\end{equation}
Here, $\ell$ is a task-specific loss function, typically the cross-entropy loss for classification tasks. In this setting, the global model is jointly trained using the clients' decentralized data and then uniformly deployed for inference on all clients.

However, such a one-size-fits-all approach often underperforms in heterogeneous environments where client data distributions vary significantly. To address this issue, \textit{pFL} aims to learn a tailored model $\theta_k$ for each client $k$. The typical objective is formulated as:
\begin{equation}\label{eq:pFL}
\min_{\{\theta_1, \ldots, \theta_N\}} \left\{ f_p := \frac{1}{N} \sum_{k=1}^N f_k(\theta_k) + \mathcal{R}(\theta_1, \ldots, \theta_N) \right\},
\end{equation}
where $\mathcal{R}$ is a regularization term designed to prevent each personalized model $\theta_k$ from overfitting to the limited local data of client~\cite{chen2022bridging}. Unlike generic FL, pFL outputs a distinct model for each client, which is used exclusively during local inference.

\subsection{Vision Transformer}
A standard ViT~\cite{dosovitskiy2020image} consists of $L$ Transformer layers. The input image is first divided into $M$ fixed-sized patches $\{a^{j}\}_{j=1}^{M}$, where each patch $a^j \in \mathbb{R}^{3 \times h \times w}$ represents a small region of the image. Each patch is then encoded into a $d$-dimensional latent embedding with positional information using a patch embedding module \texttt{Embed}:
\begin{equation}
\begin{aligned}
    x & = [a^{1}, a^{2}, \ldots, a^{M}], \quad a^j \in \mathbb{R}^{3\times h \times w}, \\
    \mathbf{E}_{0} & = [\mathbf{e}_{0}^{1}, \mathbf{e}_{0}^{2}, \ldots, \mathbf{e}_{0}^{M}], \quad \mathbf{e}_{0}^{j} = \texttt{Embed}(a^{j}) \in \mathbb{R}^{d}, 
\end{aligned}
\end{equation}
where $h$ and $w$ denote the height and width of each image patch, respectively. 
Let $\mathbf{c}_i$ and $\mathbf{E}_i$ denote the embeddings of the \texttt{[CLS]} token and the patch tokens at the input of the $(i{+}1)$-th Transformer layer $T_i$. The forward propagation in ViT proceeds as:
\begin{equation}
\begin{aligned}
    \left[ \mathbf{c}_{i}, \mathbf{E}_{i} \right] & = T_{i}([\mathbf{c}_{i-1}, \mathbf{E}_{i-1}]), & i = 1, 2, \ldots L, \\
    \hat{y} & = \texttt{Head}(\mathbf{c}_{L}),
\end{aligned}
\end{equation}
where \texttt{Head} is a classification layer that maps the \texttt{[CLS]} embedding from the final  Transformer layer, $\mathbf{c}_{L}$, to the predicted class label $\hat{y}$. 
For more details, please refer to the work~\cite{dosovitskiy2020image}.

Integrating ViT into FL offers two key advantages: (1) it helps mitigate the negative impact of data heterogeneity, and (2) it accelerates convergence by leveraging strong pre-trained representations~\cite{nguyen2022begin}. However, fine-tuning the entire ViT in FL introduces substantial computational and communication overhead. To address this, recent prompt-based tuning methods, such as VPT~\cite{jia2022visual,yang2023efficient}, enable clients to update and exchange only a small set of trainable prompts. This significantly reduces resource consumption while preserving performance. 

\subsection{Visual Prompt Tuning} 
Fully fine-tuning a ViT model typically incurs substantial computational and memory costs. To address this, inspired by prompt tuning in large language models~\cite{lester2021power}, VPT~\cite{jia2022visual} introduces a small number of trainable parameters into the input space while keeping the backbone frozen. VPT has two variants, VPT-Shallow and VPT-Deep, depending on the number of Transformer layers to which prompts are applied. A prompt is denoted as $\mathbf{p} \in \mathbb{R}^{K\times d}$, where each prompt token is a learnable $d$-dimensional vector, and $K$ is the number of prompt tokens. 
In VPT-Shallow, the prompt is inserted only into the first Transformer layer. 
The shallow-prompted ViT can be formulated as: 
\begin{equation}
\begin{aligned}
    \left[ \mathbf{c}_{1}, \mathbf{H}_{1}, \mathbf{E}_{1} \right] & = T_{1}(\left[ \mathbf{c}_{0}, \mathbf{p}, \mathbf{E}_{0} \right]), \\
    \left[ \mathbf{c}_{i}, \mathbf{H}_{i}, \mathbf{E}_{i} \right] & = T_{i}(\left[ \mathbf{c}_{i-1}, \mathbf{H}_{i-1}, \mathbf{E}_{i-1} \right]), & i = 2, 3, \ldots, L, \\
    \hat{y} & = \texttt{Head}(\mathbf{c}_{L}),
\end{aligned}
\end{equation}
where $\mathbf{H}_i$ denotes the prompt-related embeddings in the output of the $i$-th Transformer layer $T_i$. 
During fine-tuning for downstream tasks, only the prompt $\mathbf{p}$ and the classification head are updated, while the ViT backbone remains fixed. 
In contrast, VPT-Deep injects prompts at the input of every Transformer layer~\cite{jia2022visual}.
For more details, please refer to the work~\cite{jia2022visual}. 

\subsection{Semi-Implicit Variational Inference}
Given observations $\mathbf{D}$ and corresponding latent variables $\mathbf{Z}$, vanilla variational inference (VI) derives the evidence lower bound as follows: 
\begin{equation}
    \text{ELBO} = - \mathbb{E}_{\mathbf{Z}\sim q(\mathbf{Z}|\psi)} \left[\log q(\mathbf{Z}|\psi) - \log p(\mathbf{D}, \mathbf{Z})\right],
\end{equation}
where $q(\mathbf{Z}|\psi)$ is the variational distribution parameterized by $\psi$, and $p(\mathbf{D}, \mathbf{Z})$ denotes the joint distribution of the data and latent variables. 
However, VI assumes that the variational family belongs to the exponential family, which can be overly restrictive. 
To address this limitation, SIVI~\cite{yin2018semi} treats the variational parameters $\psi$ as random variables drawn from a mixing distribution. 
Specifically, in SIVI, the distribution of $\mathbf{Z}$ is defined hierarchically: $\mathbf{Z} \sim q(\mathbf{Z}|\psi)$ and $\psi \sim q_{\phi}(\psi)$, where $\phi$ parameterizes the mixing distribution $q_{\phi}(\psi)$. By marginalizing out $\psi$, one obtains a distribution family $\mathcal{H}$ over $\mathbf{Z}$:
\begin{equation}
    \mathcal{H} = \left\{ h_\phi(\mathbf{Z}): h_\phi(\mathbf{Z}) = \int_{\psi} q(\mathbf{Z}|\psi)q_{\phi}(\psi)\,d\psi \right\}.
\end{equation}

Note that $q(\mathbf{Z}|\psi)$ must be explicit, while the mixing distribution $q_{\phi}(\psi)$ can be implicit. 
The marginal distribution $h(\mathbf{Z})$ is typically implicit unless $q_{\phi}(\psi)$ is conjugate to $q(\mathbf{Z}|\psi)$. 
This combination of explicit conditionals and implicit marginals gives rise to the term \textit{semi-implicit} VI. To ensure tractability, SIVI requires that $q(\mathbf{Z}|\psi)$ be either reparameterizable~\cite{kingma2013auto} or permit an analytic ELBO. In our method, we implement $q_{\phi}(\psi)$ by randomly masking image features and transforming them through neural networks.

%% file: 4_method.tex
\section{Methodology}
In this section, we present the complete methodology of our proposed approach. We begin by deriving the variational lower bound under the SIVI framework tailored for Bayesian prompt tuning. We then describe the process of generating instance-wise visual prompts in detail. Next, we outline the optimization strategy. Finally, we summarize the inference procedure used at test time.

\subsection{Variational Lower Bound}
\label{sec:variational_lower_bound}
In VI, given the observations $\mathbf{x}$ and $\mathbf{y}$, the true posterior of the prompt $\mathbf{p}$, $p(\mathbf{p}|\mathbf{x},\mathbf{y})$, is approximated by a variational distribution $q(\mathbf{p}|\psi)$, where $\psi$ represents the variational parameters. 
To model more complex posteriors that extend beyond the exponential family, we adopt the hierarchical variational framework in SIVI and assume the following:
\begin{equation}
\mathbf{p} \sim q(\mathbf{p}|\psi), \quad \psi \sim q_{\phi}(\psi),
\end{equation} 
where $\phi$ represents the parameters of the mixing distribution.
\input{appendix/derivation}

\subsection{Prompt Generation}
Inspired by VPT-Deep~\cite{jia2022visual}, we introduce prompts into each Transformer layer of a frozen ViT model. The generation of instance-wise personalized prompts follows a two-stage sampling process. First, an intermediate latent variable $\psi$ is sampled from a conditional distribution $q_\phi(\psi|\mathbf{x})$ given the input instance $\mathbf{x}$; then, the final prompt $\mathbf{p}$ is drawn from a conditional distribution $q(\mathbf{p}|\psi)$. 

Unlike conventional variational inference methods that treat $\psi$ as a deterministic parameter, we model it as a random variable conditioned on the input features. To extract these features, we employ a frozen, pre-trained ViT model. For each input sample $\mathbf{x}$, we denote its feature representation as:
\begin{equation}
\mathbf{F} = \left\{ \mathbf{f}_1, \ldots, \mathbf{f}_L \right\},
\end{equation}
where $\mathbf{f}_i \in \mathbb{R}^{(M+1) \times d}$ represents the input token embeddings of the $i$-th layer, including the \texttt{[CLS]} token and $M$ patch tokens.

To simplify the modeling process, we assume the conditional distribution $q(\mathbf{p}|\psi)$ follows an isotropic multivariate Gaussian, i.e., $q(\mathbf{p}|\psi) = \mathcal{N}(\mathbf{p}|\mu, \Sigma)$, where the latent variable is defined as $\psi = [\mu, \Sigma]$.
To model $\mu$ and $\Sigma$ as stochastic functions of input features, we introduce randomness by applying binary masks to $\mathbf{F}$. Specifically, we first sample a set of masks:
\begin{equation}
\mathbf{M} = \left\{ \mathbf{m}_1, \ldots, \mathbf{m}_L \right\}, \quad \mathbf{m}_i \sim \texttt{Bern}(\pi)^{M+1},
\label{eq:sample_mask}
\end{equation}
and obtain masked features:
\begin{equation}
\hat{\mathbf{F}} = \left\{\hat{\mathbf{f}}_1, \ldots, \hat{\mathbf{f}}_L\right\}, \quad \hat{\mathbf{f}}_i = \mathbf{f}_i \odot \mathbf{m}_i,
\label{eq:masked_feature}
\end{equation}
where the mask element corresponding to the \texttt{[CLS]} token is always set to 1 to preserve global context.

To parameterize the distribution $q_\phi(\psi|x)$, we design a modular encoder $\mathcal{G} = \left\{ \mathbf{G}_1, \ldots, \mathbf{G}_L \right\}$, where each module $\mathbf{G}_i$ independently computes the mean and standard deviation $\left[\mu_i, \Sigma_i\right]$ for the $i$-th Transformer layer based on the masked features $\hat{\mathbf{f}}_i$:
\begin{equation}
\begin{aligned}
\left[\mu_i, \Sigma_i\right] &= \mathbf{G}_i(\hat{\mathbf{f}}_i), \quad \mu_i, \Sigma_i \in \mathbb{R}^{\nu \times d}, \quad i = 1, \dots, L,
\end{aligned}
\label{eq:mu_sigma_generation_1}
\end{equation}
where $\nu$ is the prompt length per layer. Each encoder module $\mathbf{G}_i$ consists of a Layer Normalization (LN)~\cite{ba2016layer} followed by two independent MLPs for predicting the mean and standard deviation, respectively:
\begin{equation}
\begin{aligned}
    \mu_i &= (\texttt{MLP}^{\mu}(\texttt{LN}(\hat{\mathbf{f}}_i)^{\top}))^{\top}, \\
    \Sigma_i &= \exp \left (\frac{1}{2} \cdot(\texttt{MLP}^{\Sigma}(\texttt{LN}(\hat{\mathbf{f}}_i)^\top))^\top \right),
\end{aligned}
\end{equation}
where the exponential function ensures the standard deviation remains positive, and $^\top$ denotes the transpose operation.
Since $\hat{\mathbf{f}}_i$ is a random variable conditioned on $\mathbf{f}_i$, the resulting $\mu_i$ and $\Sigma_i$ are also treated as sample-dependent random variables.
The overall prompt distribution parameters $\mu$ and $\Sigma$ are constructed by stacking the layer-wise components $\left\{\mu_i\right\}_{i=1}^{L}$ and $\left\{\Sigma_i\right\}_{i=1}^{L}$ across all $L$ layers:
\begin{equation}
\label{eq:mu_sigma_generation_2}
\begin{aligned}
\mu &= \left [ \mu_1; \dots; \mu_L \right ] \in \mathbb{R}^{L \times \nu \times d}, \\
\Sigma &= \left [ \Sigma_1; \dots; \Sigma_L \right ] \in \mathbb{R}^{L \times \nu \times d}.
\end{aligned}
\end{equation}

Using the reparameterization trick~\cite{kingma2013auto}, the prompt $\mathbf{p}$ is sampled as:
\begin{equation}
\mathbf{p} = \mu + \Sigma \odot \epsilon, \quad \epsilon \sim \mathcal{N}(0, I),
\label{eq:sampling}
\end{equation}
where $\epsilon$ is a noise tensor of the same shape as $\mu$, with elements independently drawn from the standard normal distribution. The resulting $\mathbf{p} \in \mathbb{R}^{L \times \nu \times d}$ serves as the final instance-wise prompt.

To further improve the stability of federated training, we introduce an additional global prompt $\bar{\mathbf{p}}$ that is shared across all instances, following prior works~\cite{yang2023efficient,feng2023learning,deng2024unlocking}. The final prompt input for each sample is formed by concatenating the global prompt with the instance-specific personalized prompt, i.e., $\mathbf{p}_{\text{total}} = [\bar{\mathbf{p}}, \mathbf{p}]$. Accordingly, the model prediction is conditioned on both components: $p(\mathbf{y} | \bar{\mathbf{p}}, \mathbf{p}, \mathbf{x})$. For notational simplicity, we use $\mathbf{p}$ to denote the concatenated prompt throughout the remainder of the paper.

\subsection{Optimization}
In Section~\ref{sec:variational_lower_bound}, we have derived the lower bound $\mathcal{\underline{L}}$ for the ELBO. 
However, directly optimizing $\mathcal{\underline{L}}$ may lead to a degeneracy issue, where $q_{\phi}(\psi)$ collapses to a point mass density. 
This would cause SIVI to reduce to the vanilla VI~\cite{yin2018semi}. 
To prevent degeneracy, we follow the work~\cite{yin2018semi} and introduce the following regularization term:
\begin{equation}
\begin{aligned}
\mathcal{A}_{S} &= \mathbb{E}_{\psi, \tilde{\psi}^{1},\ldots, \tilde{\psi}^{S}} \text{KL}(q(\mathbf{p}|\psi) \Vert \tilde{h}_{S}(\mathbf{p})),
\end{aligned}
\end{equation}
where
\begin{equation}
\begin{aligned}
\tilde{h}_{S}(\mathbf{p}) = \frac{q(\mathbf{p}|\psi) + \sum_{s=1}^{S}q(\mathbf{p}|\tilde{\psi}^{s})}{S+1}. 
\end{aligned}
\end{equation}
This regularization term $\mathcal{A}_{S}$ satisfies two properties: (1) $\mathcal{A}_{S} \geq 0$; (2) $\mathcal{A}_{S} = 0$ if and only if $S = 0$ or $q_{\phi}(\psi)$ degenerates to a point mass density. 
According to the work~\cite{yin2018semi}, the augmented objective $\mathcal{\underline{L}}_{S} = \mathcal{\underline{L}} + \mathcal{A}_{S}$ serves as an asymptotically exact surrogate ELBO that $\mathcal{\underline{L}}_{0} = \mathcal{\underline{L}}$ and $\lim_{S \to \infty}\mathcal{\underline{L}}_S = \mathcal{L}$. 
Maximizing $\mathcal{\underline{L}}_{S}$ with $S \geq 1$ derives positive $\mathcal{A}_{S}$ and could drive $q_{\phi}(\psi)$ away from degeneracy. 
Moreover, importance reweighting~\cite{burda2015importance} can be further introduced to tighten $\mathcal{\underline{L}}_{S}$ by drawing $J$ samples $\{(\mathbf{p}^{j}, \psi^{j})\}_{j=1}^{J}$ from $q(\mathbf{p}, \psi|\mathbf{x})$. 
The final objective can be formulated as
\begin{equation}
\label{eq:final_objective}
\begin{aligned}
\underline{\mathcal{L}}_{S}^{J} = \mathbb{E}_{\left\{(\mathbf{p}^{j}, \psi^{j})\right\}_{j=1}^J \sim q(\mathbf{p}|\psi)q_{\phi}(\psi|\mathbf{x})} \mathbb{E}_{\left\{\tilde{\psi}^{s}\right\}_{s=1}^S \sim q_{\phi}(\psi|\mathbf{x})} 
\\ 
\left[\text{log}~\frac{1}{J}\sum_{j=1}^{J}\frac{p(\mathbf{y}, \mathbf{p}^{j}|\mathbf{x})}{\Omega^j}
\right],
\end{aligned}
\end{equation}
where 
\begin{equation}
    \begin{aligned}
        \Omega^j = \frac{1}{S+1}\left[q(\mathbf{p}^{j}|\psi^{j}) + \sum_{s=1}^{S}q(\mathbf{p}^{j}|\tilde{\psi}^{s})\right].
    \end{aligned}
\end{equation}

For simplicity, the above objective is defined with respect to a single training sample $(\mathbf{x}, \mathbf{y})$. In practice, each client performs optimization over its entire local dataset $\mathcal{D}$. The overall training objective is thus defined as the average over all local samples:
\begin{equation}
\label{eq:average_final_objective}
\mathcal{J} = \frac{1}{|\mathcal{D}|} \sum_{(\mathbf{x}, \mathbf{y}) \in \mathcal{D}} \underline{\mathcal{L}}_{S}^{J}(\mathbf{x}, \mathbf{y}),
\end{equation}
where $\underline{\mathcal{L}}_{S}^{J}(\mathbf{x}, \mathbf{y})$ denotes the objective for sample $(\mathbf{x}, \mathbf{y})$. The expectation can be approximated using mini-batch sampling, and the objective $\mathcal{J}$ is optimized via stochastic gradient ascent.

\input{algorithm/algorithm}

\subsection{Inference}
To enhance the model’s capacity for capturing image features, we sample $V$ prompt vectors $\left\{\mathbf{p}^v\right\}_{v=1}^V$ from the prompt posterior for each test instance. Specifically, we first draw $V$ binary masks $\left\{\mathbf{M}^v\right\}_{v=1}^V$ from a Bernoulli distribution \texttt{Bern($\pi$)}, and apply them to the image feature representation $\mathbf{F}$ to obtain masked features. These are then used to generate $V$ sets of variational parameters $\left\{\psi^{v} := \left[\mu^{v}, \Sigma^{v}\right]\right\}_{v=1}^{V}$. 
During inference, we bypass stochastic sampling from $q(\mathbf{p}^{v}|\psi^v)$ and instead set each prompt vector $\mathbf{p}^v$ to the corresponding mean $\mu^v$.
These prompts yield $V$ predicted class distributions $\left\{p(\mathbf{y}|\mathbf{p}^{v}, \mathbf{x})\right\}_{v=1}^{V}$, and the final prediction is obtained by averaging the probabilities and selecting the class with the highest score:
\begin{equation}
\hat{y} = \arg\max_{c} \left\{ p(\mathbf{y} = c | \mathbf{x}) := \frac{1}{V} \sum_{v=1}^{V} p(\mathbf{y} = c | \mathbf{p}^v, \mathbf{x}) \right\}.
\label{eq:test_argmax}
\end{equation}

\subsection{Summary}
During each round of federated training, each client first performs multiple local optimization steps to maximize the objective $\mathcal{J}$ defined in Equation~\ref{eq:average_final_objective}.
After local updates, clients synchronize with the server by uploading the updated parameters, including the global prompts $\bar{\mathbf{p}}$, the classifier parameters $\theta_{\texttt{Head}}$, and the encoder parameters $\phi$.
The server then aggregates these parameters across clients.
The complete training procedure of \algo~ is summarized in Algorithm~\ref{alg:pfedbayespt}.

%% file: appendix/derivation.tex
We marginalize $\psi$ out and derive
\begin{equation}
    \mathbf{p} \sim h_\phi(\mathbf{p}) = \int_{\psi} q(\mathbf{p}|\psi)q_{\phi}(\psi)\,d\psi.
\end{equation}
We maximize the log-likelihood of the observations $\mathbf{y}$ conditioned on $\mathbf{x}$, and apply Jensen's inequality~\cite{jensen1906fonctions} to derive: 
\begin{equation}
\begin{aligned}
& \log p(\mathbf{y}|\mathbf{x}) \\
&= \log \int p(\mathbf{y}, \mathbf{p}|\mathbf{x}) \,d\mathbf{p} \\
&= \log \int \frac{p(\mathbf{y}, \mathbf{p}|\mathbf{x})}{h_{\phi}(\mathbf{p}|\mathbf{x})} h_{\phi}(\mathbf{p}|\mathbf{x}) \,d\mathbf{p} \\
&\geq \mathbb{E}_{h_{\phi}(\mathbf{p}|\mathbf{x})}\left[\log \frac{p(\mathbf{y}, \mathbf{p}|\mathbf{x})}{h_{\phi}(\mathbf{p}|\mathbf{x})}\right] = \mathcal{L},
\end{aligned}
\end{equation}
where $\mathcal{L}$ is the ELBO and 
\begin{equation}
\begin{aligned}
h_{\phi}(\mathbf{p}|\mathbf{x}) = \int q(\mathbf{p}|\psi) q_{\phi}(\psi|\mathbf{x}) \,d\psi
\end{aligned}
\end{equation}
is the marginal distribution over $\mathbf{p}$. 
Then a lower bound of the ELBO can be derived: 
\begin{equation}
\begin{aligned}
\mathcal{L} &= \mathbb{E}_{h_{\phi}(\mathbf{p}|\mathbf{x})}\left[\log\frac{p(\mathbf{y}, \mathbf{p}|\mathbf{x})}{h_{\phi}(\mathbf{p}|\mathbf{x})}\right] \\
   &= \mathbb{E}_{h_{\phi}(\mathbf{p}|\mathbf{x})}\left[\log \frac{p(\mathbf{p}|\mathbf{x}, \mathbf{y})p(\mathbf{y}|\mathbf{x})}{h_{\phi}(\mathbf{p}|\mathbf{x})}\right] \\
   &= -{\text{KL}}\left(h_{\phi}\left(\mathbf{p}|\mathbf{x}\right) \Vert p\left(\mathbf{p}|\mathbf{x},\mathbf{y}\right)\right) + \log p(\mathbf{y}|\mathbf{x}) \\
   &= - {\text{KL}} \left(\mathbb{E}_{\psi \sim q_\phi\left(\psi|\mathbf{x}\right)}q(\mathbf{p}|\psi) \Vert p\left(\mathbf{p}|\mathbf{x},\mathbf{y}\right)\right) + \log p(\mathbf{y}|\mathbf{x}) \\
   &\geq -\mathbb{E}_{\psi \sim q_{\phi}(\psi|\mathbf{x})} {\text{KL}}(q(\mathbf{p}|\psi) \Vert p(\mathbf{p}|\mathbf{x},\mathbf{y})) + \log p(\mathbf{y}|\mathbf{x}) \\
   &= \mathbb{E}_{\psi \sim q_{\phi}(\psi|\mathbf{x})} \mathbb{E}_{\mathbf{p} \sim q(\mathbf{p}|\psi)} \left[\log \frac{p(\mathbf{y}, \mathbf{p}|\mathbf{x})}{q(\mathbf{p}|\psi)}\right] \\ 
   &= \mathcal{\underline{L}},
\end{aligned}
\end{equation}
where ${\text{KL}}$ represents the Kullback-Leibler divergence and we employ ${\text{KL}}(\mathbb{E}_{\psi \sim q_\phi(\psi)}q(\mathbf{p}|\psi) \Vert p(\mathbf{p})) \leq \mathbb{E}_{\psi \sim q_{\phi}(\psi)}{\text{KL}}(q(\mathbf{p}|\psi) \Vert p(\mathbf{p}))$ according to SIVI~\cite{yin2018semi}. 

To better understand $\mathcal{\underline{L}}$, we further expand it as:
\begin{equation}
\label{eq:generator_classifier}
\begin{aligned}
\mathcal{\underline{L}} &= \mathbb{E}_{\psi \sim q_{\phi}(\psi|\mathbf{x})} \mathbb{E}_{\mathbf{p} \sim q(\mathbf{p}|\psi)} \left[\log \frac{p(\mathbf{y}, \mathbf{p}|\mathbf{x})}{q(\mathbf{p}|\psi)}\right] \\
      &= \mathbb{E}_{\psi \sim q_{\phi}(\psi|\mathbf{x})} \mathbb{E}_{\mathbf{p} \sim q(\mathbf{p}|\psi)} \left[\log \frac{p(\mathbf{y}|\mathbf{p}, \mathbf{x})p(\mathbf{p}|\mathbf{x})}{q(\mathbf{p}|\psi)}\right] \\
      &= \mathbb{E}_{\psi \sim q_{\phi}(\psi|\mathbf{x})} \mathbb{E}_{\mathbf{p} \sim q(\mathbf{p}|\psi)} \left[\log p(\mathbf{y}|\mathbf{p}, \mathbf{x})\right] \\ 
      &\quad 
      - \mathbb{E}_{\psi \sim q_{\phi}(\psi|\mathbf{x})}{\text{KL}}(q(\mathbf{p}|\psi) \Vert p(\mathbf{p}|\mathbf{x})).
\end{aligned}
\end{equation}
Here, $\psi \sim q_\phi(\psi|\mathbf{x})$ and $\mathbf{p} \sim q(\mathbf{p}|\psi)$ denote the personalized prompt generation process, which is modeled via neural networks. The term $p(\mathbf{y}|\mathbf{p}, \mathbf{x})$ corresponds to the classifier that outputs the predictive distribution conditioned on the input $\mathbf{x}$ and its associated prompt $\mathbf{p}$.
Equation~\ref{eq:generator_classifier} clearly reflects the principle of semi-implicit Bayesian optimization:

\begin{itemize}
\item The first term represents the log-likelihood, measuring the model’s predictive capacity for the label $\mathbf{y}$ with the aid of the generated personalized prompt $\mathbf{p}$; this can be interpreted as the negative classification loss.
\item The second term is a KL divergence regularization, which constrains the prompt distribution $q(\mathbf{p}|\psi)$ to remain close to the prior distribution $p(\mathbf{p}|\mathbf{x})$, thereby mitigating overfitting and enhancing generalization.
\end{itemize}

%% file: algorithm/algorithm.tex
\begin{algorithm}
\caption{Training of \algo~}
\label{alg:pfedbayespt}
\SetKwInOut{KwData}{Input}
\SetKwInOut{KwResult}{Output}
\KwData{
Initialized global prompt $\bar{\mathbf{p}}$, 
parameters of the classification head $\theta_{\texttt{Head}}$, 
parameters of the encoder $\phi$, 
number of communication rounds $R$, 
number of local update epochs $E$, 
local mini-batch size $B$, 
learning rate for $\bar{\mathbf{p}}$ and $\theta_{\texttt{Head}}$: $\eta$, 
learning rate for $\phi$: $\rho$, number of samples for KL regularization $S$, 
number of importance-weighted samples $J$, and Bernoulli parameter $\pi$
}
\KwResult{
Final global prompt $\bar{\mathbf{p}}$, encoder parameters $\phi$, and classifier parameters $\theta_{\texttt{Head}}$
}
\For{each round $r=1$ to $R$} {
    Server selects a subset of clients $\mathcal{S}^r$ \\
    \For{each client $c_{k} \in \mathcal{S}^r$}{
	\For{each epoch $e=1$ to $E$}{
        Split local dataset $\mathcal{D}_k$ into mini-batches $\mathcal{B}_k$ of size $B$ \\
		\For{each mini-batch $(\mathbf{X}, \mathbf{Y}) \in \mathcal{B}_{k}$}{
                \For{each sample $(\mathbf{x}, \mathbf{y}) \in (\mathbf{X}, \mathbf{Y})$ (in parallel)}{
                    Sample $\left\{\tilde{\psi}^{s}\right\}_{s=1}^{S}$ according to Eq.~\ref{eq:mu_sigma_generation_1} and~\ref{eq:mu_sigma_generation_2} \\
                    Sample $\left\{( \mathbf{p}^j,\psi^j) \right\}_{j=1}^{J}$ according to Eq.~\ref{eq:mu_sigma_generation_1}, ~\ref{eq:mu_sigma_generation_2} and ~\ref{eq:sampling} \\
                    Compute $\underline{\mathcal{L}}_{S}^{J}(\mathbf{x}, \mathbf{y})$ according to Eq.~\ref{eq:final_objective} \\
                }
                Compute the mini-batch objective $\mathcal{J}$ using Eq.~\ref{eq:average_final_objective} \\
                $\bar{\mathbf{p}} \leftarrow \bar{\mathbf{p}} + \eta \nabla_{\bar{\mathbf{p}}} \mathcal{J}$ \\
		      $\theta_{\texttt{Head}} \leftarrow \theta_{\texttt{Head}} + \eta  \nabla_{\theta_{\texttt{Head}}} \mathcal{J}$ \\
		      $\phi \leftarrow \phi + \rho \nabla_{\phi} \mathcal{J}$ \\
            }
	}
    }
    Server aggregates $\bar{\mathbf{p}}$, $\theta_{\texttt{Head}}$, and $\phi$ from selected clients, updates the global model, and synchronizes it to all clients \\
}
\end{algorithm}

%% file: 5_results.tex
\input{tables/domainnet}
\section{Experiments}
In this section, we present comprehensive experiments to evaluate the effectiveness of \algo~. We aim to answer the following research questions:
\begin{itemize}
    \item \textbf{(RQ1)} How does \algo~ perform under different types of data heterogeneity?
    \item \textbf{(RQ2)} Can \algo~ generalize effectively to clients not involved in training?
    \item \textbf{(RQ3)} How does the number of prompt samples $V$ during inference affect performance?
    \item \textbf{(RQ4)} Do the Bayesian modeling and implicit posterior play a critical role in improving performance?
\end{itemize}

\subsection{Experimental Setup}
\subsubsection{Datasets}
We evaluate the effectiveness of our method under two distinct types of data heterogeneity: feature shift and label shift. These correspond to variations in the conditional distribution $p(\mathbf{x}|\mathbf{y})$ and the label distribution $p(\mathbf{y})$ across clients, respectively.

To simulate feature shift, we use the DomainNet dataset, which contains approximately 0.6 million images spanning 345 categories, distributed across six domains: Clipart, Infograph, Painting, Quickdraw, Real, and Sketch.
Following the work~\cite{yang2023efficient}, we construct a sub-dataset using the ten most frequent classes.\footnote{To ensure balance, we subsample each domain to match the size of the smallest domain, resulting in 1,830 samples per domain.}
We then partition the data across 6 clients, with each client assigned $m$ domains. In our experiments, $m$ is varied from 1 to 6 to simulate different degrees of feature shift. For the label shift setting, we partition the CIFAR-100 dataset into 100 clients, each containing at most $s$ classes. We vary $s$ in $\left\{5, 10, 15, 20, 50\right\}$ to simulate different levels of heterogeneity in label distribution.

\subsubsection{Baselines} We compare our proposed method against the following competitive baselines:
\begin{itemize}
    \item \textbf{Head-Tune}~\cite{sun2022exploring}: All clients collaboratively tune the global classification head using FedAvg~\cite{mcmahan2017communication}, while keeping the backbone frozen.
    \item \textbf{FedVPT}~\cite{jia2022visual}: FedVPT inserts visual prompts into the input layer and optimizes global prompts using FedAvg~\cite{mcmahan2017communication}. Following the implementation in the work~\cite{yang2023efficient}, each client transmits only the visual prompts to the server in each communication round, while retaining its local classification head.
    \item \textbf{FedVPT-D}~\cite{jia2022visual}: An extension of FedVPT that prepends prompts to both the input and all hidden layers of the model.
    \item \textbf{pFedPG}~\cite{yang2023efficient}: Based on hypernetworks~\cite{ha2017hypernetworks,shamsian2021personalized}, pFedPG trains a server-side prompt generator to produce client-specific prompts conditioned on client descriptors.
    \item \textbf{FedPR}~\cite{feng2023learning}: Originally proposed for MRI reconstruction, FedPR is adapted here for classification by augmenting FedVPT-D with a null space projection step in the local update of visual prompts.
    \item \textbf{SGPT}~\cite{deng2024unlocking}: A state-of-the-art federated visual prompt tuning method that jointly trains shared prompts and a prompt selection module, enabling the model to capture both common and group-specific features.
\end{itemize}

\subsubsection{Implementation details}
Following the work~\cite{yang2023efficient}, we use ViT-B/16~\cite{dosovitskiy2020image} as the frozen backbone and set the prompt length to 10. 
The personalized prompt length per layer is fixed to $\nu = 1$. For DomainNet, all clients participate in each communication round. For CIFAR-100, 5\% of clients are randomly sampled in each round. 
We set the number of local epochs $E=5$ and total communication rounds $R=100$ for all experiments. 
For FedVPT and FedVPT-D, the local learning rate is set to 0.01. 
For pFedPG, we tune the local learning rate from $\{0.01, 0.05, 0.1, 0.25\}$ and the prompt generator's learning rate from $\{0.001, 0.01, 0.05, 0.1\}$. 
For FedPR, following the work~\cite{feng2023learning}, we use the last 80\% of the singular values to construct the null space of the global prompt. We implement both FedPR-Shallow and FedPR-Deep, corresponding to VPT-Shallow and VPT-Deep~\cite{jia2022visual}, and report the better-performing variant. 
For SGPT, to ensure fair comparison with FedVPT, each client is assigned a personalized local classification head. 

We implement \algo~ in PyTorch and train it using the SGD optimizer. 
We set $\pi=0.9$, and tune the insertion depth of global and instance-wise prompts from layers 1 to 12, respectively (starting from the first Transformer block).
To reduce computational and memory overhead during training, we fix $S = J = 1$ in Equation~\ref{eq:final_objective}.
During inference, we set the number of sampled prompts to $V = 5$ in Equation~\ref{eq:test_argmax}.
The learning rate $\eta$ for both the classification head and global prompts is set to 0.01, consistent with FedVPT and FedVPT-D. The learning rate $\rho$ for the encoder is tuned over $\left\{0.0001, 0.0005, 0.001, 0.005, 0.01\right\}$.

We evaluate all methods using two metrics: (1) \textit{Average}, which denotes the average test accuracy across all clients and (2) \textit{Worst Local}, whihc denotes the lowest test accuracy among all clients. 
Following the work~\cite{deng2024unlocking}, we report the average performance over the last 10 communication rounds for each metric. All experiments are conducted on a single NVIDIA A800 GPU. Each experiment is repeated three times using random seeds 0, 1, and 2, and we report the mean across the three runs.

\input{tables/cifar100}

\subsection{Main Results (RQ1)}
We present the experimental results on DomainNet (feature heterogeneity) in Table~\ref{tab:domainnet} and on CIFAR-100 dataset (label heterogeneity) in Table~\ref{tab:cifar100}.
\subsubsection{Feature Heterogeneity}
Across all settings, \algo~ consistently outperforms all baselines under both the \textit{Average} and \textit{Worst Local} metrics.

Focusing first on the \textit{Average} metric: Head-Tune serves as a generic federated learning approach that trains a single global classifier shared across all clients. As $m$ increases, meaning each client covers more domains and inter-client heterogeneity decreases, Head-Tune achieves improved performance. The other methods introduce different levels of personalization to enhance accuracy. Specifically, FedVPT, FedVPT-D, and FedPR adopt personalized classification heads; pFedPG further generates client-specific prompts based on learned descriptors; SGPT introduces group-level prompts for individual instances. When $m$ is small, the data within each client becomes more homogeneous, making local optimization easier and amplifying the benefits of personalization.

Specifically, pFedPG generates personalized prompts for each client and consistently outperforms FedVPT across all settings. FedVPT-D further enhances performance by applying prompts to both the input and hidden layers, enabling the model to capture more comprehensive global representations, and therefore generally surpasses FedVPT. SGPT achieves stronger results by providing finer-grained, group-level prompts for each instance. Leveraging its instance-wise prompt generation and Bayesian optimization framework, \algo~ consistently achieves the best performance across all settings. Although its accuracy slightly declines as $m$ increases, the performance margin between \algo~ and the strongest baseline remains notable. In all configurations, \algo~ maintains an advantage of approximately 1\% over the second-best method, SGPT.

For the \textit{Worst Local} metric, a consistent decline in performance is observed across all methods as $m$ decreases. This trend is largely due to the presence of particularly challenging domains in DomainNet, such as Infograph and Quickdraw. When $m$ is small, these difficult domains may dominate the local data of certain clients, leading to poor performance on those outlier clients and thus lowering the overall worst-case score. Notably, \algo~ achieves the highest \textit{Worst Local} accuracy in all cases, highlighting its robustness to skewed client distributions.

\subsubsection{Label Heterogeneity}
Turning to the \textit{Average} metric under the label shift scenario, we observe a similar pattern to that seen in feature shift. As $s$ increases, the label distribution across clients becomes more balanced, leading to improved performance for Head-Tune. Conversely, when $s$ is small, each client contains fewer classes, which simplifies the classification task and favors methods with stronger personalization capabilities. 

However, unlike in the feature shift case, FedVPT-D performs worse than FedVPT in most configurations. This is because the deeper prompts used in FedVPT-D are located closer to the model’s decision layers and tend to encode task-specific semantics. Under severe label heterogeneity, such shared prompts can introduce negative transfer between clients, reducing the generalization ability of the global model.

In this setting, pFedPG still maintains a consistent advantage over FedVPT. Across all configurations, \algo~ consistently outperforms the best-performing baseline, SGPT. A similar trend holds for the \textit{Worst Local} metric, further demonstrating the effectiveness of the proposed method.

\input{tables/generalization}

\subsection{Generalization to New Clients (RQ2)}
In pFL, personalization typically refers to tailoring models for clients involved in the training process. However, most existing pFL methods struggle to provide effective personalization for new clients that were not seen during training—especially when there exists a substantial distributional shift between new and training clients. This limitation reduces the practical applicability of such methods. In contrast, our proposed method generates instance-wise personalized prompts, enabling it to generalize effectively to unseen clients. To evaluate the generalization capability of \algo~ under various data heterogeneity conditions, we conduct the following experiments.

Specifically, we simulate six clients with $m \in \{1, 2\}$ on DomainNet and one hundred clients with $s \in \{5, 10\}$ on CIFAR-100. For each dataset, half of the clients are used for training, and the remaining half are treated as unseen clients during training. The participation ratio is set to 1 for DomainNet and 0.1 for CIFAR-100. Before inference, each new client independently fine-tunes its local classification head for five epochs. For pFedPG, following the work~\cite{shamsian2021personalized}, we also update the client descriptor to generate personalized prompts for the new clients. In this setting, we focus exclusively on evaluating performance on the unseen clients.

The results are reported in Table~\ref{tab:generalization}. On DomainNet, all methods perform substantially better when $m=2$ than when $m=1$, as the training clients cover more domains. On CIFAR-100, clients with fewer local classes ($s=5$) yield better performance due to the relative simplicity of the classification task. pFedPG demonstrates moderate improvements over FedVPT, FedVPT-D, and FedPR across most configurations. SGPT achieves strong generalization performance, benefitting from its prompt selection mechanism. Nevertheless, \algo~ consistently outperforms all baselines across different settings, surpassing the best-performing baseline, SGPT, by approximately 1\%.

\subsection{Effect of the Number of  Sampled Prompts (RQ3)}
\begin{figure}[htbp!]
\centering
\includegraphics[width=\linewidth]{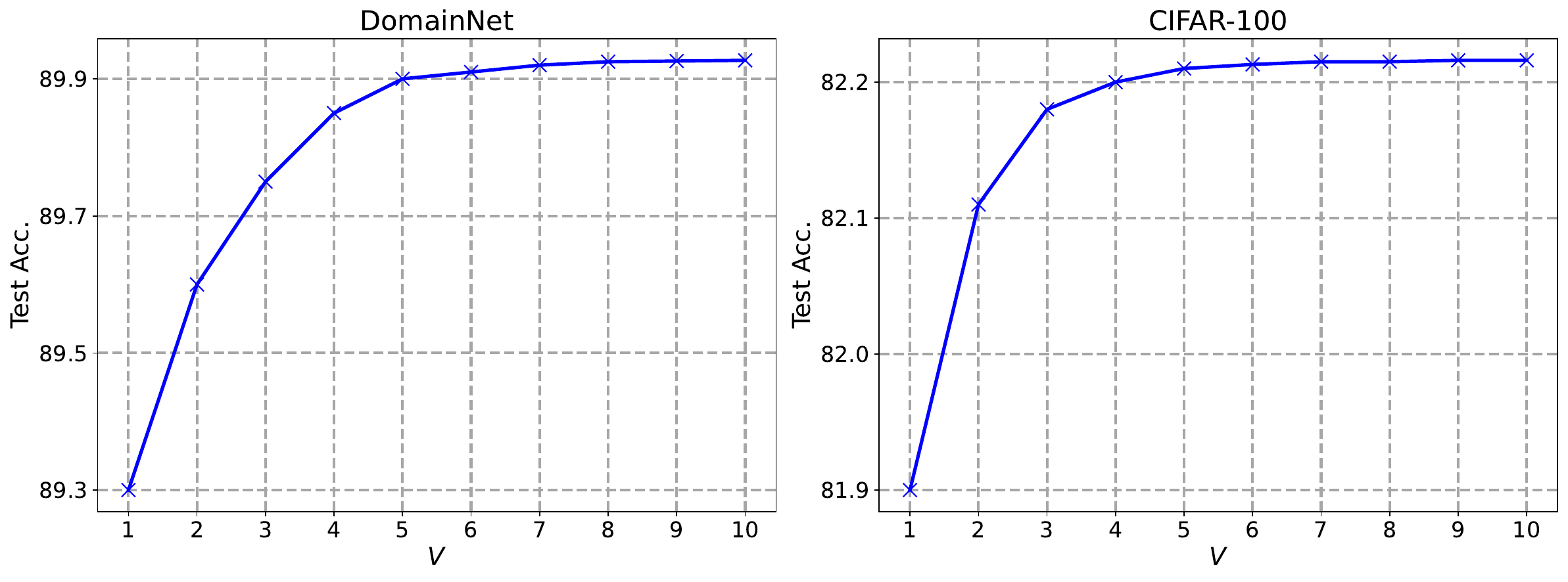}
\caption{Average test accuracy of \algo~ under different numbers of sampled prompts ($V$).}
\label{fig:pfedbayespt_varying_V}
\end{figure}

A key advantage of \algo~ is its ability to learn a posterior distribution over instance-wise prompts, enabling multiple samples to be drawn during inference to improve prediction accuracy. However, increasing the number of sampled prompts also leads to higher computational cost. 
To strike a balance between performance and efficiency, we evaluate how varying the number of samples affects model performance. 
Experiments are conducted on DomainNet ($m=6$) and CIFAR-100 ($s=50$), with the number of sampled prompts $V \in \{1, 2, \dots, 10\}$. For each setting, we report the average test accuracy (\textit{Average}) across all clients. The results are presented in Figure~\ref{fig:pfedbayespt_varying_V}.

As shown in the figure, test accuracy improves consistently with an increasing number of prompt samples. On DomainNet, accuracy peaks at 89.90\% when $V=5$, consistent with the results reported in Table~\ref{tab:domainnet}. Further increasing $V$ yields only marginal gains. A similar trend is observed on CIFAR-100, where accuracy reaches 82.21\% at $V=5$ (see Table~\ref{tab:cifar100}) and shows limited improvement beyond this point.

These results confirm the effectiveness of our prompt sampling mechanism: even a few samples can substantially improve accuracy. 
In federated settings, clients with more computational resources can further enhance performance by increasing the number of sampled prompts, making \algo~ both practical and scalable by balancing predictive accuracy with inference cost.

\subsection{Ablation Studies (RQ4)}
Table~\ref{tab:ablation_studies} presents an ablation study comparing different variants of \algo~ to assess the contribution of each component.
In particular, \algo~-G models the prompt posterior as a standard Gaussian distribution, while \algo~-D treats the prompt as a deterministic vector.
\begin{table}[htbp!]
\centering
\caption{Comparison of different \algo~ variants.}
\label{tab:ablation_studies}
\resizebox{1.0\linewidth}{!}{
\begin{tabular}{cccc}
\toprule
Method      & pFedBayesPT-D   & pFedBayesPT-G & pFedBayesPT \\ 
\midrule
DomainNet   & 88.70     & 89.41     & 89.90 \\ 
CIFAR-100   & 81.11     & 81.65     & 82.21 \\ 
\bottomrule
\end{tabular}
}
\end{table}

As shown in the table, \algo~-D yields the lowest performance and even falls behind the strongest baseline, SGPT (see Tables~\ref{tab:domainnet} and~\ref{tab:cifar100}), likely due to its lack of uncertainty modeling, which increases the risk of overfitting.
\algo~-G shows a notable improvement over \algo~-D, confirming the benefit of introducing Bayesian modeling into prompt tuning.
Finally, the full version of our method, \algo~, further improves upon \algo~-G, highlighting the effectiveness of modeling the prompt posterior as a complex implicit distribution rather than a simple Gaussian.

%% file: tables/domainnet.tex
\begin{table*}[htbp!]
\centering
\caption{Quantitative comparisons on DomainNet. \textbf{Bold} denotes the best result.}
\label{tab:domainnet}
\begin{tabular}{lcccccccccccc}
\toprule
{Dataset} & \multicolumn{12}{c}{DomainNet} \\ 
\cmidrule(lr){1-1} \cmidrule(lr){2-13}
\multirow{2}{*}{Method}
& \multicolumn{6}{c}{Average}
& \multicolumn{6}{c}{Worst Local}
\\
\cmidrule{2-7}\cmidrule{8-13}
& $m=1$        
& $m=2$      
& $m=3$   
& $m=4$ 
& $m=5$        
& $m=6$  

& $m=1$        
& $m=2$      
& $m=3$   
& $m=4$ 
& $m=5$        
& $m=6$ 
\\ 

\midrule
Head-Tune
& 84.77 
& 84.90
& 85.03 
& 85.10
& 85.12
& 85.13

& 59.87       
& 70.01           
& 76.82
& 77.52
& 80.31
& 83.46

\\
FedVPT
& 88.30       
& 86.16        
& 85.12                           
& 84.25
& 83.76
& 83.29

& 61.09
& 71.97
& 75.44
& 76.31
& 78.54
& 81.07
\\
FedVPT-D
& 90.81      
& 90.05         
& 89.13                   
& 88.93
& 88.40
& 88.31

& 66.81
& 79.39
& 82.52
& 83.54
& 84.87
& 87.25

\\
pFedPG
& 88.41            
& 86.49             
& 85.48                 
& 84.79
& 84.07
& 83.87

& 61.91
& 72.98
& 75.85
& 75.84
& 79.58
& 81.92
\\

FedPR
& 90.60           
& 89.12          
& 88.66                    
& 87.67 
& 87.62
& 87.32

& 66.48
& 76.56
& 81.99
& 81.32
& 83.75
& 86.13
\\

SGPT
& {91.20}
& {90.55}           
& {89.66}                          
& {89.43}
& {88.87}
& {88.80}

& {67.33}
& {80.01}
& {82.93}
& {84.00}
& {85.29}
& {88.11}
\\

\midrule
\algo~  
& \textbf{92.33}       
& \textbf{91.62}           
& \textbf{90.73}                    
& \textbf{90.39} 
& \textbf{89.91} 
& \textbf{89.90}

& \textbf{68.85}
& \textbf{81.20}
& \textbf{84.21}
& \textbf{85.11}
& \textbf{86.33}
& \textbf{89.09}
\\
\bottomrule
\end{tabular}
\end{table*}

%% file: tables/cifar100.tex
\begin{table*}[htbp!]
\centering
\caption{Quantitative comparisons on CIFAR-100. \textbf{Bold} denotes the best result.}
\label{tab:cifar100}
\begin{tabular}{lcccccccccc}
\toprule
{Dataset} & \multicolumn{10}{c}{CIFAR-100} \\ 
\cmidrule(lr){1-1} \cmidrule(lr){2-11}
\multirow{2}{*}{Method}
& \multicolumn{5}{c}{Average}        
& \multicolumn{5}{c}{Worst Local}   
\\
\cmidrule(lr){2-6}
\cmidrule(lr){7-11}
& $s=5$ 
& $s=10$   
& $s=15$ 
& $s=20$ 
& $s=50$ 
& $s=5$ 
& $s=10$ 
& $s=15$ 
& $s=20$ 
& $s=50$ 
\\
\midrule
Head-Tune
& 74.86      
& 75.77
& 76.45
& 76.78
& 77.03

& 53.28
& 56.85
& 63.72
& 65.45
& 66.80
\\
FedVPT
& 95.07            
& 92.41
& 89.28
& 87.39
& 75.12

& 79.84
& 72.86
& 71.47
& 67.67
& 49.69
\\
FedVPT-D
& 95.34      
& 87.47                
& 78.83                           
& 77.41 
& 63.36

& 79.76
& 69.94
& 52.25
& 56.20
& 45.44

\\
pFedPG
& 95.79            
& 92.94             
& 90.66                           
& 89.06
& 80.51

& 81.57
& 76.52
& 74.44
& 74.39
& 67.81
\\

FedPR
& 95.82          
& 91.44                
& 88.37                           
& 81.42 
& 58.87

& 82.04
& 74.81
& 73.85
& 63.56
& 41.54
\\

SGPT
& 96.22           
& 93.50              
& 91.11                      
& 89.76
& 81.32

& 82.44
& 77.11
& 74.99
& 74.69
& 68.23
\\

\midrule
\algo~  
& \textbf{97.18}          
& \textbf{94.61}           
& \textbf{92.41}                         
& \textbf{91.11}
& \textbf{82.21}

& \textbf{83.39}
& \textbf{78.22}
& \textbf{76.07}
& \textbf{75.71}
& \textbf{69.60}
\\
\bottomrule
\end{tabular}
\end{table*}

%% file: tables/generalization.tex
\begin{table}[htbp!]
\centering
\caption{Performance comparison on unseen clients under different heterogeneity levels. \textbf{Bold} denotes the best results.}
\label{tab:generalization}
\begin{tabular}{lcccccc}
\toprule
{Dataset} 
& \multicolumn{2}{c}{DomainNet}
& \multicolumn{2}{c}{CIFAR-100} \\ 
\cmidrule(lr){1-1} \cmidrule(lr){2-3} \cmidrule(lr){4-5}
{Method}
& $m=1$
& $m=2$
& $s=5$
& $s=10$
\\
\midrule
Head-Tune
& 81.00                     
& 84.26
& 94.47
& 89.08

\\
FedVPT
& 84.81                      
& 86.27
& 95.70
& 90.32

\\
FedVPT-D
& 87.37                      
& 89.51
& 95.93
& 93.07

\\
pFedPG
& 88.71                      
& 89.07
& 96.40
& 94.40
\\

FedPR
& 86.83
& 89.45
& 96.38
& 93.77
\\

SGPT
& 89.23           
& 90.35
& 97.01              
& 95.12
\\

\midrule
\algo~  
& \textbf{90.32}         
& \textbf{91.44}     
& \textbf{98.10}                      
& \textbf{96.08}
\\
\bottomrule
\end{tabular}
\end{table}

%% file: 6_conclusion.tex
\section{Conclusion}
In this paper, we presented \algo~, a novel instance-wise personalized federated learning framework that leverages semi-implicit Bayesian prompt tuning. 
By formulating prompt generation as a variational inference problem and modeling the posterior of instance-wise prompts with an implicit distribution, our approach effectively enhances prompt diversity and expressiveness while mitigating the risk of overfitting on data-scarce clients. 
Comprehensive experiments on benchmark datasets validate that \algo~ achieves superior performance compared with state-of-the-art personalized FL baselines under both feature and label heterogeneity, demonstrating its strong capability for fine-grained personalization. 
Looking ahead, our work opens up new directions for exploring Bayesian perspectives in federated prompt learning, and we plan to extend our framework to broader modalities and more challenging real-world federated environments.